\documentclass[letterpaper, 10 pt, conference]{ieeeconf}  
\pdfoutput=1
\IEEEoverridecommandlockouts                              
\usepackage{graphicx}        
\usepackage{multicol}        
\usepackage[bottom]
           {footmisc}        
\usepackage{etoolbox}        
\usepackage[noend]{algorithmic}           
\usepackage{amsmath}                      
\usepackage{amsfonts}                     
\usepackage{amssymb}
\usepackage{amsthm}
\usepackage{mathrsfs}                     
\usepackage{float} 
\usepackage{graphicx} 
\usepackage{nicefrac}
\usepackage{ calc }           
\usepackage[usenames,dvipsnames,svgnames,table]
           {xcolor}
\usepackage{caption}
\usepackage{subcaption}
\usepackage[dvips,frame,rotate,import]{xy}                   

\title{\LARGE \bf Fast Collision Checking: From Single Robots to Multi-Robot Teams}

\author{Joshua Bialkowski,  Michael Otte, and Emilio Frazzoli
 \thanks{%
 J. Bialkowski, M. Otte, and E. Frazzoli,
 Massachusetts Institute of Technology, 
 Cambridge, MA. {\tt\small ottemw@mit.edu}
 }}
           
\theoremstyle{definition}

\theoremstyle{definition}

\makeatletter
\newcommand{\removelatexerror}{\let\@latex@error\@gobble}
\makeatother

\newlength{\tightalgowidth}
\newlength{\tightalgoremainder}

\makeatletter
\patchcmd{\@algocf@start}{%
  \begin{lrbox}{\algocf@algobox}%
}{%
  \rule{0.5\tightalgoremainder}{\z@}%
  \begin{lrbox}{\algocf@algobox}%
  \begin{minipage}{\tightalgowidth}%
}{}{}
\patchcmd{\@algocf@finish}{%
  \end{lrbox}%
}{%
  \end{minipage}%
  \end{lrbox}%
}{}{}
\makeatother

\begin{document}


%
%
\maketitle

\thispagestyle{empty}
\pagestyle{empty}

\begin{abstract} 
We examine three different algorithms that enable the collision certificate method from \cite{Bialkowski.wafr12} to handle the case of a centralized multi-robot team. By taking advantage of symmetries in the configuration space of multi-robot teams, our methods can significantly reduce the number of collision checks vs.\ both \cite{Bialkowski.wafr12} and standard collision checking implementations.
 
\end{abstract}

\section{Introduction}

Collision checking is a critical bottle-neck in robotic motion planning and a key hurtle to enabling more sophisticated real-time 
robotic systems \cite{lavalle.book06}. 
Collision checking for a multi-robot team is even more difficult than for a single robot. 
In the centralized motion planning planning problem, the hyper-volume of a configuration space scales exponentially vs.\ the number of robots in the team and is often correlated with collision checking runtime.

\begin{figure}[b]
  \hspace{.1cm}
  \begin{xy}
    \xyimport(100,100){\includegraphics[width=.94\linewidth, trim=0 120 0 0, clip=true]{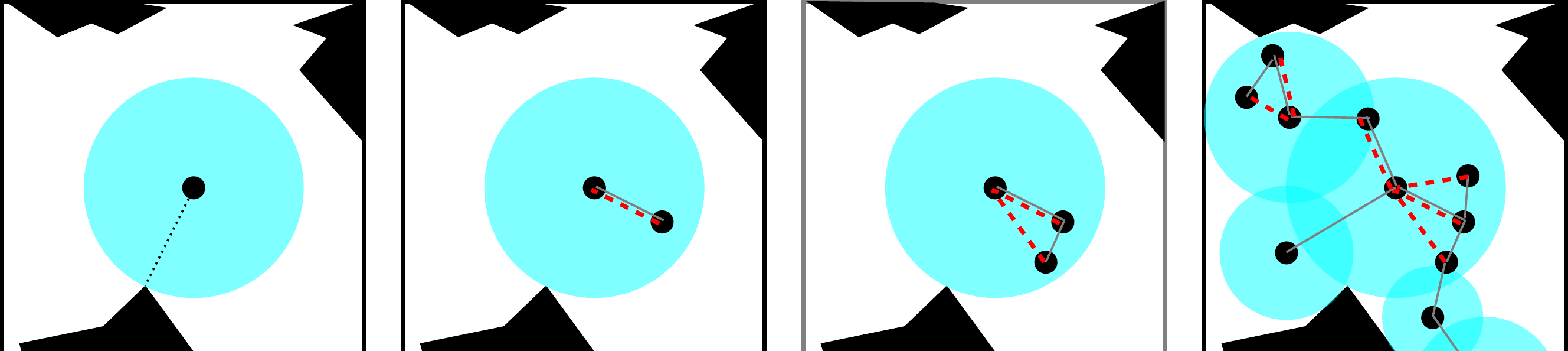}}
    ,(10,50)*{p}
    ,(16,26)*{D_{min}}
    ,(36,50)*{p}
    ,(41,26)*{q}
  \end{xy}

 \footnotesize

\hspace{1cm}(A)\hspace{1.7cm}(B)\hspace{1.7cm}(C)\hspace{1.7cm}(D)\\

 \caption[]{Our single-robot collision certificate method from \cite{Bialkowski.wafr12}.
Collision checked nodes $p$ store ``safety certificates'' (blue) defined by $D_{min}$ the distance to the nearest obstacle (A). Future nodes $q$ within a certificate can forgo collision checking (B). Pointer (red-dotted lines) are maintained to certifying nodes (C). The ratio of collision checks vs. (all) nodes approaches zero in the limit vs. graph size (D).

}

  \label{fig:single_robot}
\end{figure}

In \cite{Bialkowski.wafr12} we show that collision checking can be significantly reduced for a {\it single}-robot in a metric space by using ``safety certificates'' that record $D_{min}$, the (normally) collision-checked distance of a point $p$ to the nearest obstacle, see Figure~\ref{fig:single_robot}.
If a new node $q$ is drawn from within an existing certificate (i.e., $\| p - q \| < D_{min}$), than $q$ cannot possibly be in collision and a new check (for $q$) is {\it unnecessary}. $q$ then stores a pointer to $p$ so that future nodes drawn near $q$ can also check their status vs.\ the certificate stored at $p$.
Certificates can be stored within a kd-tree (which is already a common subroutine in motion planning algorithms, e.g., \cite{Lavalle.ijrr01, Kavraki.trans96, Karaman.ijrr11}), and so our method can be used without increasing the asymptotic runtime complexity of many common motion planning algorithms. Moreover, the expected proportion of collision checks vs. all samples approaches zero as the number of samples increases to infinity (see \cite{Bialkowski.wafr12} for details).
\textbf{We now extend this result to centralized \textit{multi}-robot teams. 
}

\section{Multi-Robot Algorithms}

Assuming that the members of a multi-robot team share an environment, then the configuration space of the multi-robot team is a Cartesian product of the space of each robot\footnote{If all members of the team do not share an environment, then it may be possible to reduce the problem by dividing it into a set of disjoint sub-problems, one per each set of robots that are common to a particular environment, and such that each robot belongs to only one team.}. $R$ robots each planning in $D$-dimensions yields a $RD$-dimensional configuration space.

In the centralized multi-robot problem, collision checking vs.\ the environment can be accomplished piecewise per robot\footnote{Robot vs. robot collision checking can similarly be reduced to checking one robot vs.\ another in a local coordinate system. If the team is homogeneous then this is further simplified because all robots can re-use the same two-robot collision checking data-structure, since the robot vs. robot interaction will be identical for any pair of robots. That said, we only address robot vs.\ environment collision checks in the current work. However, we note that the current work can be applied to a two-robot collision check (and thus by extension the homogeneous team self-collision check) by recasting a robot vs. robot check as a robot vs.\ environment check, where the second environment simply consists of a single robot.}.
In the current paper we evaluate three safety certificate methods for multi-robot teams that we call \textit{Basic Certificate}, \textit{Partial Certificate}, and \textit{Shared Projection}. We now describe the individual algorithms.

\subsection{Basic Certificate}

\begin{figure}[t]

  \begin{xy}
    \xyimport(100,100){\includegraphics[width=\linewidth, trim=0 0 0 0, clip=true]{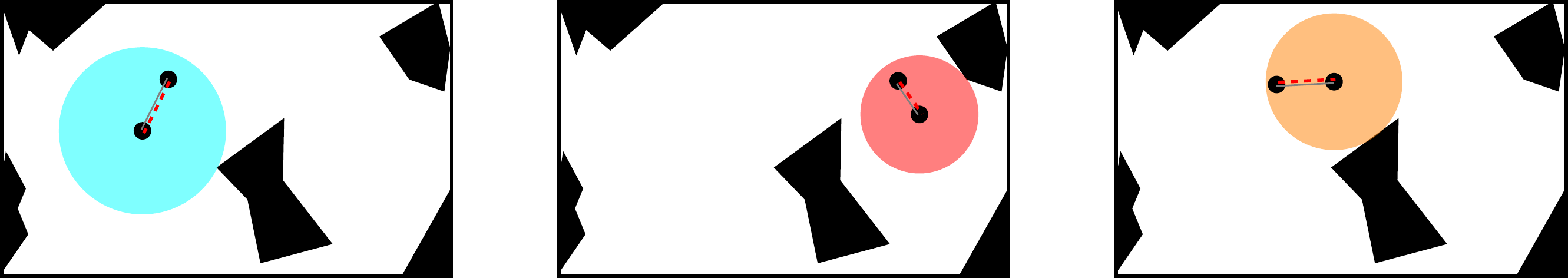}}
    ,(32,50)*{+}
    ,(68,50)*{+}
    ,(4,11)*{A}
    ,(40,11)*{B}
    ,(76,11)*{C}
    ,(12,50)*{p_a}
    ,(14,70)*{q_a}
    ,(61.5,53)*{p_b}
    ,(55,70)*{q_b}
    ,(88,74)*{p_c}
    ,(79,74)*{q_c}
  \end{xy}

\caption[]{\textbf{\textit{Basic Certificate}}: Certificates are a Cartesian product of balls, one ball per robot. $q$ is certified safe by $p$ if the projections $q_a$, $q_b$ and $q_c$ are in the projected certificates of $p_a$, $p_b$ and $p_c$ (blue, red, and orange balls), respectively.}
  \label{fig:shared}


\end{figure}

In the {\it Basic Certificate} method, certificates are a Cartesian product of $D$-balls such that there is one $D$-ball per robot. Consider the case where three robots share a $2$-dimensional workspace, see Figure~\ref{fig:shared}. The robots are labeled $A$, $B$, and $C$, respectively, and the subscripts $a$, $b$, and $c$ denote a particular robot's projection of a point. If, e.g., the team is located at point $p$ within the combined configuration space, then robot $a$ is located at the projected point $p_a$ within its own projection of that space. The certificate stored at $p$ is defined to be $[D_{min,a}, D_{min,b}, D_{min,c}]$, a list of the radii of its three balls.

Point $q$ is certified safe by $p$ if the projections $q_a$, $q_b$ and $q_c$ are in the projected certificates of $p_a$, $p_b$ and $p_c$ (blue, red, and orange balls), respectively. In other words, $q$ is certified safe by $p$ if $\|p_i-q_i\| < D_{min,i}$ for $i \in \{a, b, c\}$.

Note that {\it Basic Certificate} is exactly the method presented in \cite{Bialkowski.wafr12} applied to a centralized multi-robot team {\it as is}, and does not make any special considerations for symmetries in the space. If any robot is outside its own projection of the certificate, then the {\it entire} team is deemed to be outside the certificate and a new collision check must be performed.
The projected space of each robot must be {\it independently} covered with certificates---despite the fact that each robot has to face an identical obstacle configuration.  
The next two methods are designed to address these limitations.

\subsection{Partial Certificate}

\begin{figure}[t]
  \begin{xy}
    \xyimport(100,100){\includegraphics[width=\linewidth, trim=0 0 0 0, clip=true]{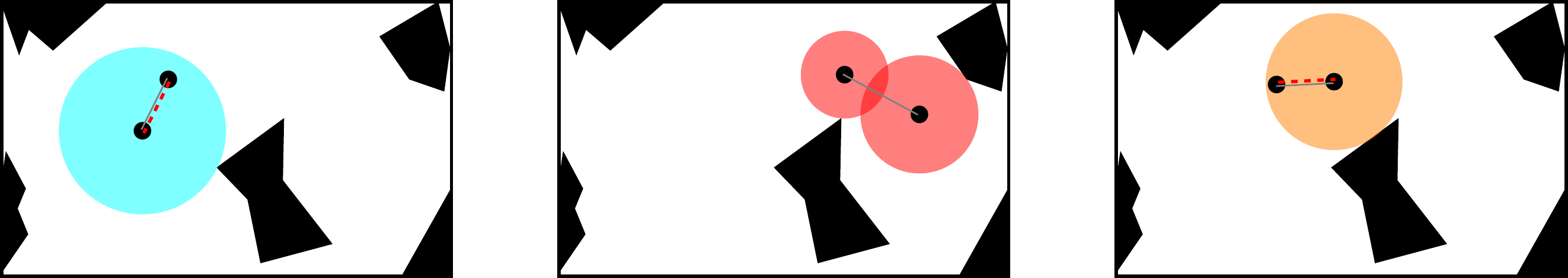}}
    ,(32,50)*{+}
    ,(68,50)*{+}
    ,(4,11)*{A}
    ,(40,11)*{B}
    ,(76,11)*{C}
    ,(12,50)*{p_a}
    ,(14,70)*{q_a}
    ,(61.5,53)*{p_b}
    ,(51.5,74)*{q_b}
    ,(88,74)*{p_c}
    ,(79,74)*{q_c}
  \end{xy}
  \caption[]{\textbf{\textit{Partial Certificate}}: If a point is {\it not} certified as safe with respect to a subspace projection, then only a partial collision check is required. e.g, $q_a$ and $q_c$ are within the certificates of $p_a$ and $p_c$, respectively, but $q_b$ is not within the certificate of $p_b$. Thus, only $1/3$ check is required (for $q_b$).}
  \label{fig:partial}
\end{figure}

In {\it Partial Certificate} each robot performs collision checking in its own projection of the full space (similar to {\it Basic Certificate}). 
However, if a point is {\it not} certified as safe with respect to all subspace projection, then only a partial collision check is required vs.\ the projection(s) that were not individually certified as safe. See Figure~\ref{fig:partial}, when $q_a$ and $q_c$ are within the certificates of $p_a$ and $p_c$, respectively, but $q_b$ is not within the certificate of $p_b$, then only $1/3$ check is required to determine the safety of $q_b$. 

The implementation of this method requires that each node stores $R$ certificate pointers (i.e., instead of the single pointer required by {\it basic certificate}). Storing one pointer per robot enables a new node to be certified by a combination of different old nodes and/or to calculate its own partial certificates as needed.

\subsection{Shared Projection}

\begin{figure}[t]
  \begin{xy}
    \xyimport(100,100){\includegraphics[width=\linewidth, trim=0 0 0 0, clip=true]{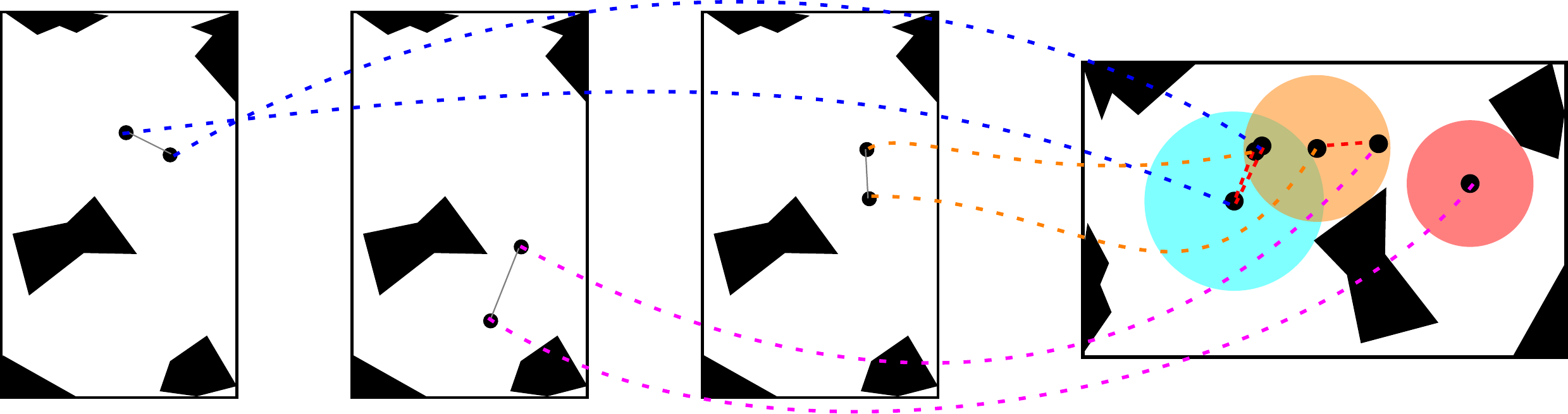}}
    ,(19,50)*{+}
    ,(41,50)*{+}
    ,(2,20)*{A}
    ,(25,20)*{B}
    ,(48,20)*{C}
    ,(5.5,65)*{p_a}
    ,(13,54)*{q_a}
    ,(29.5,29)*{p_b}
    ,(35.5,46)*{q_b}
    ,(56,42)*{p_c}
    ,(52.5,66)*{q_c}
  \end{xy}

  \footnotesize

\caption[]{\textbf{\textit{Shared Projection}}: all robots collision check in the {\it same} $D$-dimensional projection (far right). Pointers from configuration space node projections to their collision-checking projection counterparts are depicted with blue/magenta/orange dotted lines, respectively. Note that $p_a$ certifies $q_c$ and $p_c$ certifies $q_b$.}
  \label{fig:shared}

\end{figure}

\textit{Shared Projection} is similar to {\it Partial Certificate} except that all robots perform collision checking in the {\it same} $D$-dimensional projection of the full $RD$-dimensional space, See Figure~\ref{fig:shared}. We anticipate that doing this will cause the latter space to become populated with certificates $R$ times more quickly, and thus require fewer (standard) collision checks. In other words, vs.\ a single projection of the space, $R$ nodes are checked and/or added during each iteration instead of $1$. Therefore, we expect that it will require $1/R$ iterations to achieve the same amount of certified space vs.\ {\it Partial Certificate} (and {\it Basic Certificate}).

The implementation of this method requires an {\it extra} $D$-dimensional kd-tree in the shared projected space; however, the time complexity only increases from $\mathcal{O}(RD\log(N))$ to ${\mathcal{O}( RD\log(N) + D\log(R) )}$, where $D\log(R)$ is a constant.

In practice, we find that the runtime of this method can be significantly improved by seeding the second (shared $D$-dimensional) kd-tree based on the nearest-neighbor as determined by of the first ($(RD)$-dimensional) kd-tree. For example, when searching the second (shared $D$-dimensional) kd-tree for the collision status of $q_a$ we begin the search at the location of $p_a$ instead of at the root of the tree, where $p$ is the point that has (already) been returned by the first ($(RD)$-dimensional) kd-tree search.

\section{experiments}

\begin{figure}[t]

  \footnotesize

 \includegraphics[width=\linewidth, trim=100 230 100 230, clip=true]{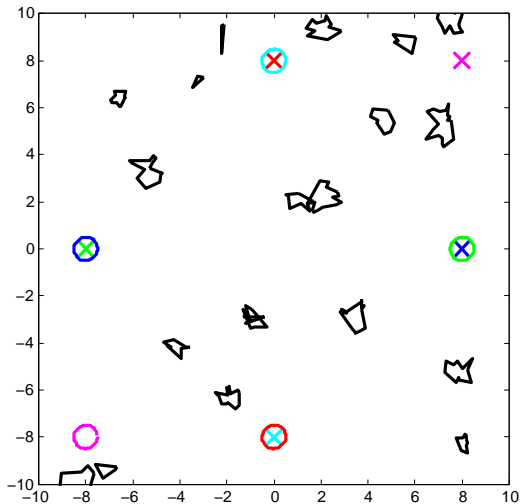}

   \caption[]{The randomly generated workspace that is used for all experiments. Robot starting locations and goals appear as circles and crosses, respectively. Note that a particular robot's starting location and goal have the same color. Obstacle appear black and have been randomly generated.}
    \label{fig:environment}
\end{figure}

We now perform a number of experiments evaluating the effectiveness of using {\it Basic Certificate}, {\it Partial Certificate}, and {\it Shared Projection} with RRT and RRT*, and vs.\ different obstacle checking times and team sizes. The workspace used for all experiments appears in Figure~\ref{fig:environment}, note that robots $1$, $2$, $3$, $4$, and $5$ are colored blue, red, green, cyan, and magenta, respectively. When an experiment is run with a team size of $R$ then robots numbered $1-R$ are used and robots $\geq R+1$ are removed from the workspace.

\begin{figure}[h]

\centering

\footnotesize
 
RRT

\vspace{.3cm}

\includegraphics[width=\linewidth, trim=40 312 95 315, clip=true]{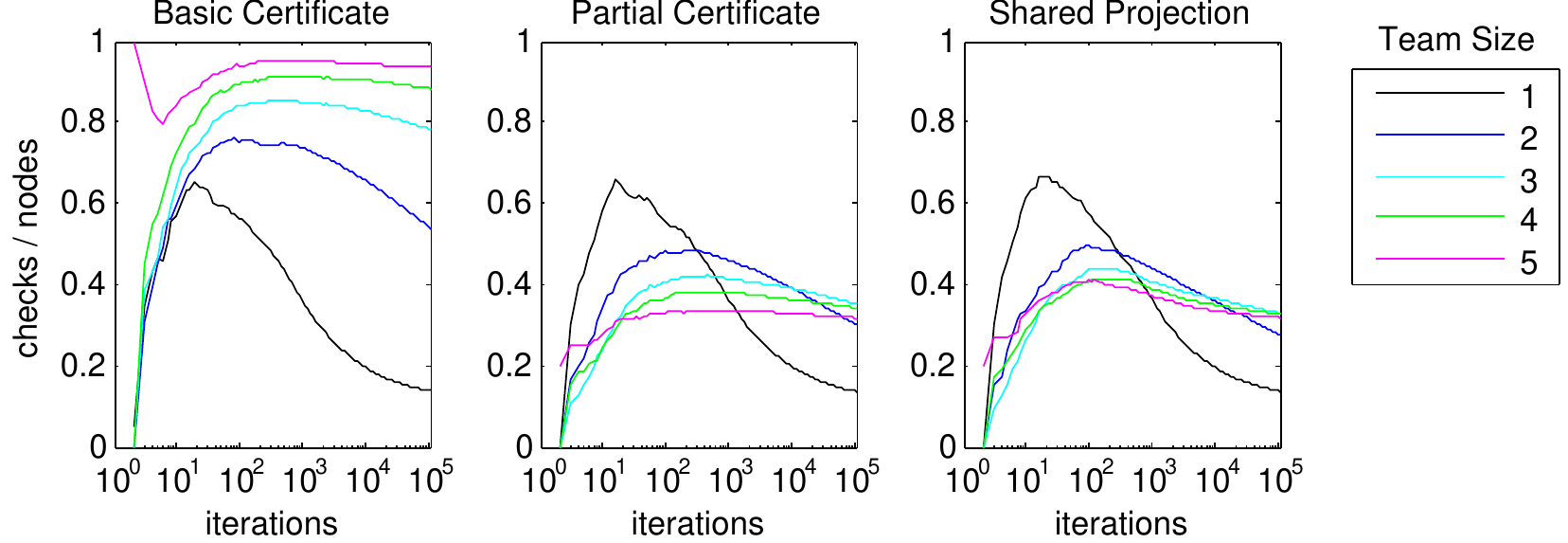}

   \caption[]{Proportion of nodes requiring a collision check (mean value over 20 trials), lower values are better. The RRT algorithm is used.}
    \label{fig:ratio}

\end{figure}

\begin{figure}[h]

  \centering

  \footnotesize

 RRT*

\vspace{.3cm}

 \includegraphics[width=\linewidth, trim=40 312 95 315, clip=true]{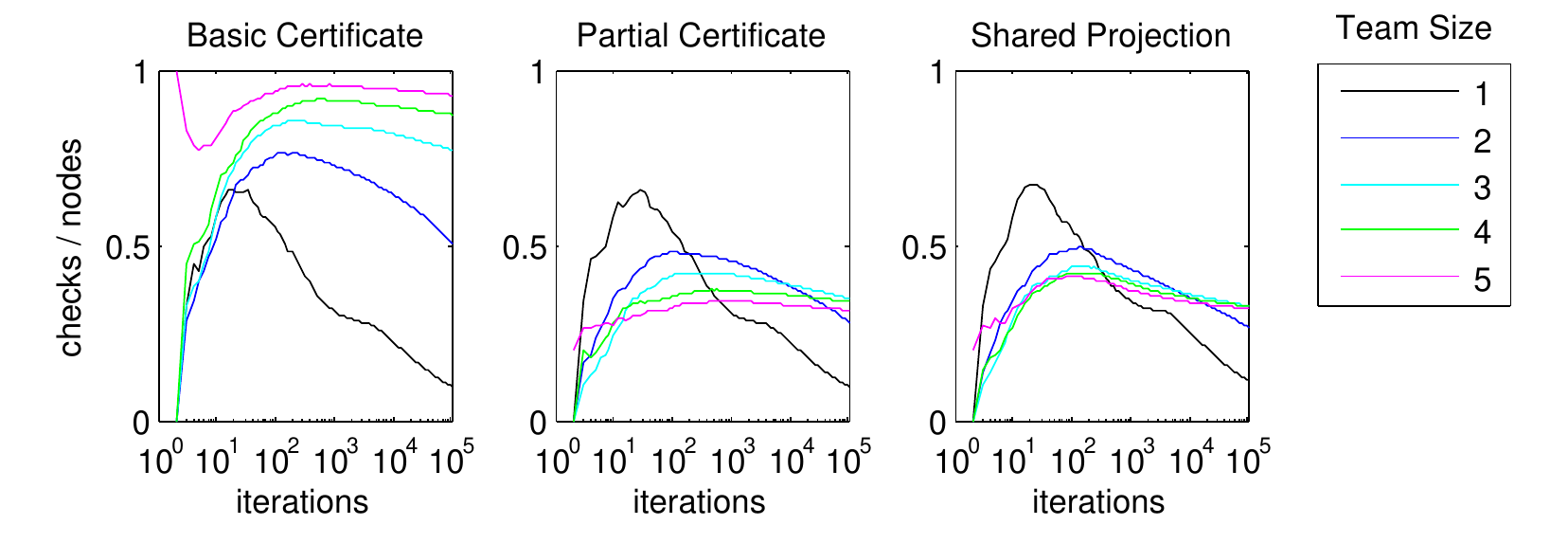}

   \caption[]{Proportion of nodes requiring a collision check (mean value over 20 trials), lower values are better. The RRT* algorithm is used.}
    \label{fig:ratio_star}
\end{figure}

Figures~\ref{fig:ratio}~and~\ref{fig:ratio_star} show the average proportion of points that require a collision check (over 20 trials) for different team sizes (1 to 5 robots) vs.\ iteration number ($1$ to $10^5$). Note that fractional values are possible for {\it Partial Certificate} and {\it Shared Projection} when only some of the robots require a check. Figure~\ref{fig:ratio} shows results with RRT, while \ref{fig:ratio_star} shows results with RRT*.

\begin{figure}[h]

\centering

\footnotesize
 
RRT

\vspace{.3cm}

 \includegraphics[width=\linewidth, trim=40 312 95 315, clip=true]{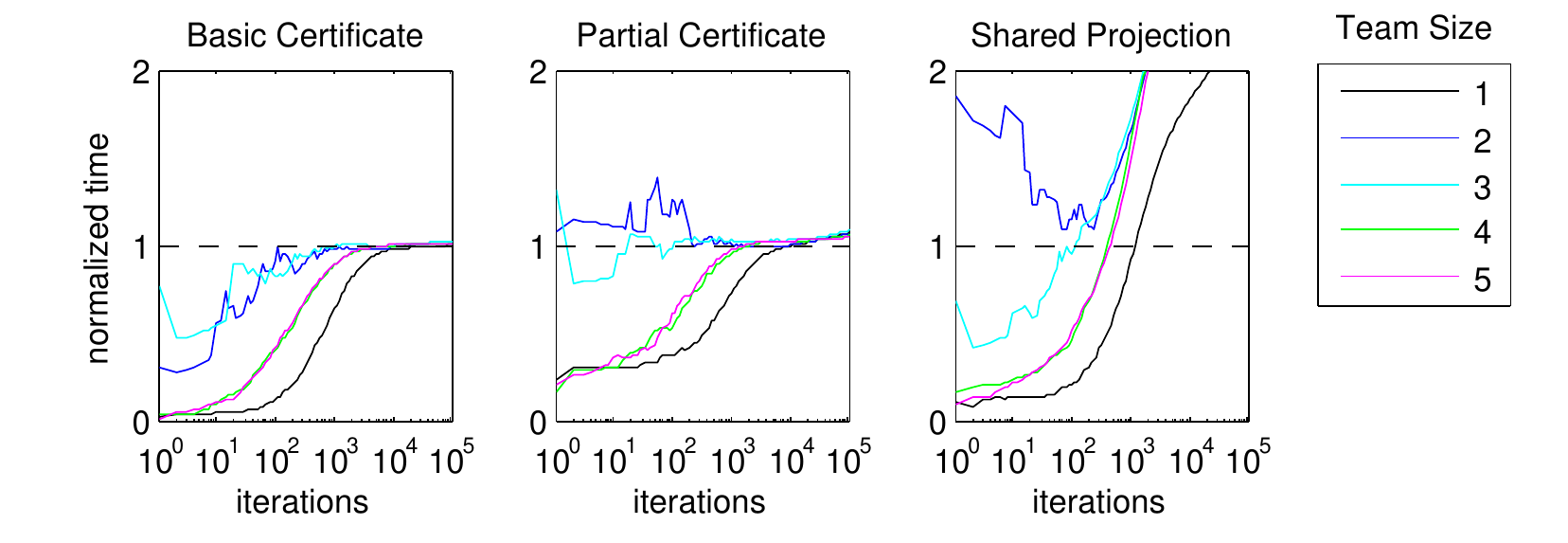}

\vspace{.3cm}

RRT (Difficulty $10^2$)

\vspace{.3cm}

 \includegraphics[width=\linewidth, trim=40 312 95 315, clip=true]{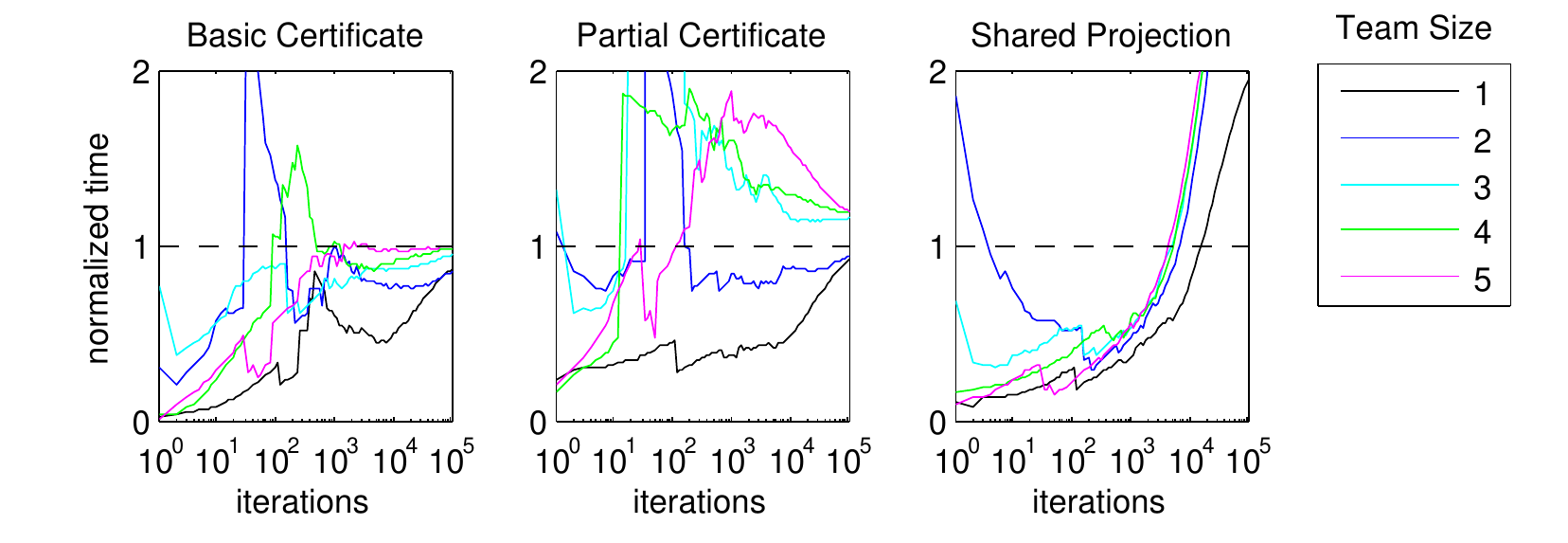}

\vspace{.3cm}

RRT (Difficulty $10^4$)

\vspace{.3cm}

 \includegraphics[width=\linewidth, trim=40 312 95 315, clip=true]{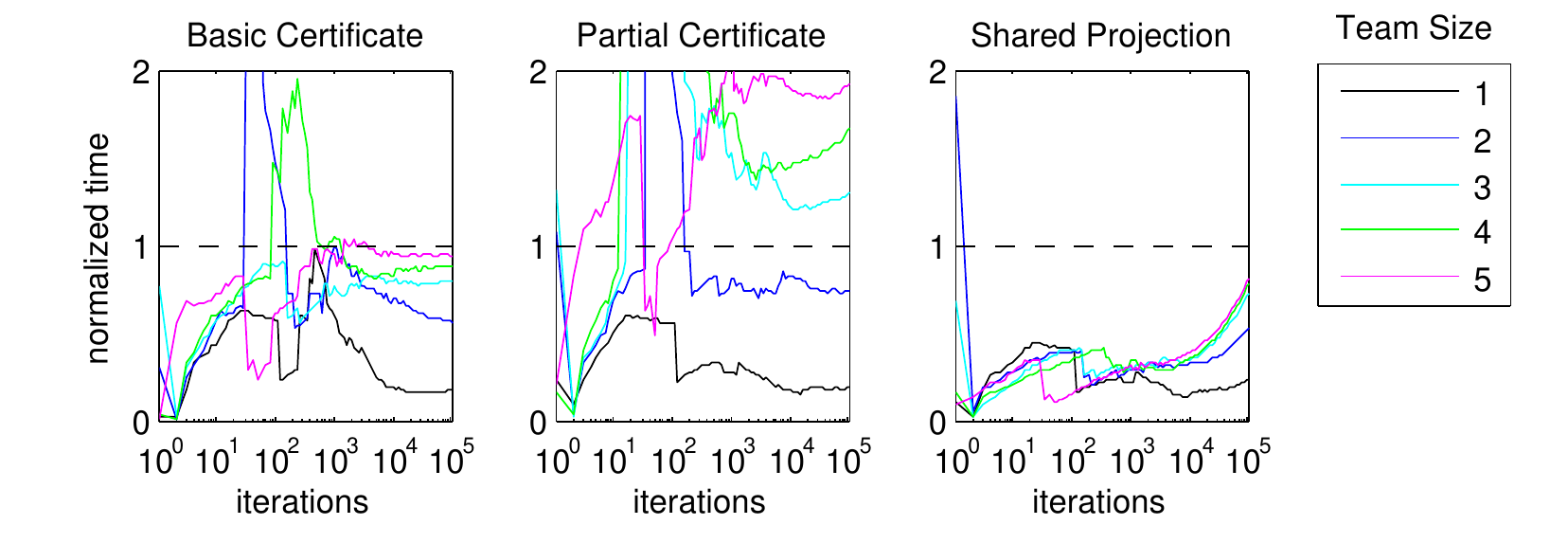}

    \caption[]{Relative runtime of certificate methods vs. normal collision checking (mean over 20 trials), points below the dotted line are desired. The RRT algorithm is used. Top to Bottom depict the RRT algorithm with collision difficulty levels of $1$ (raw data), $10^2$, and $10^4$, respectively.}
    \label{fig:time}

\end{figure}

\begin{figure}[h]

\centering

\footnotesize
 
RRT*

\vspace{.3cm}

\includegraphics[width=\linewidth, trim=40 312 95 315, clip=true]{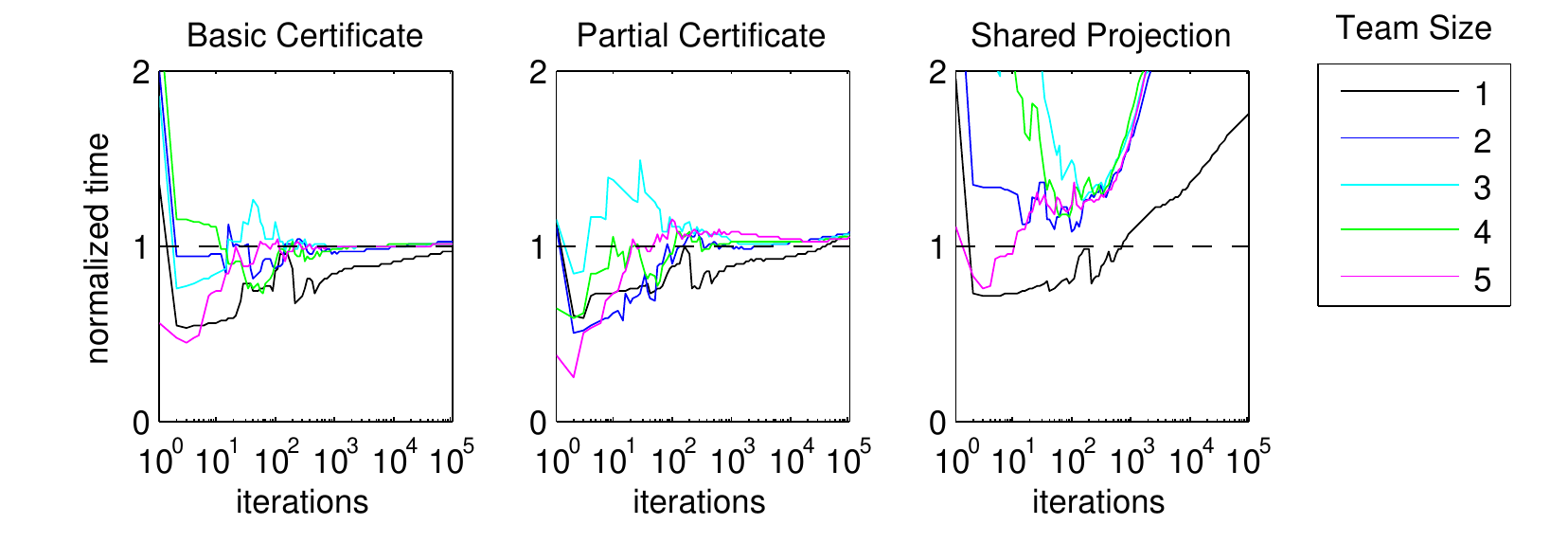}

\vspace{.3cm}

RRT* (Difficulty $10^2$)

\vspace{.3cm}

\includegraphics[width=\linewidth, trim=40 312 95 315, clip=true]{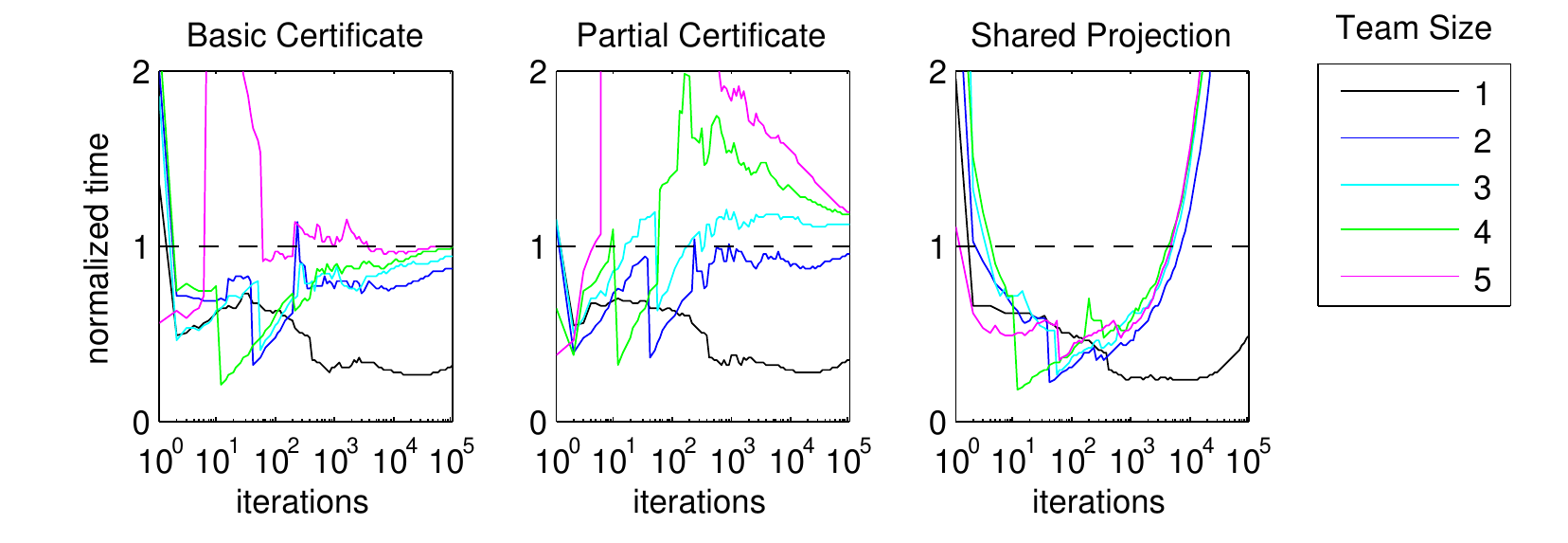}

\vspace{.3cm}

RRT* (Difficulty $10^4$)

\vspace{.3cm}

\includegraphics[width=\linewidth, trim=40 312 95 315, clip=true]{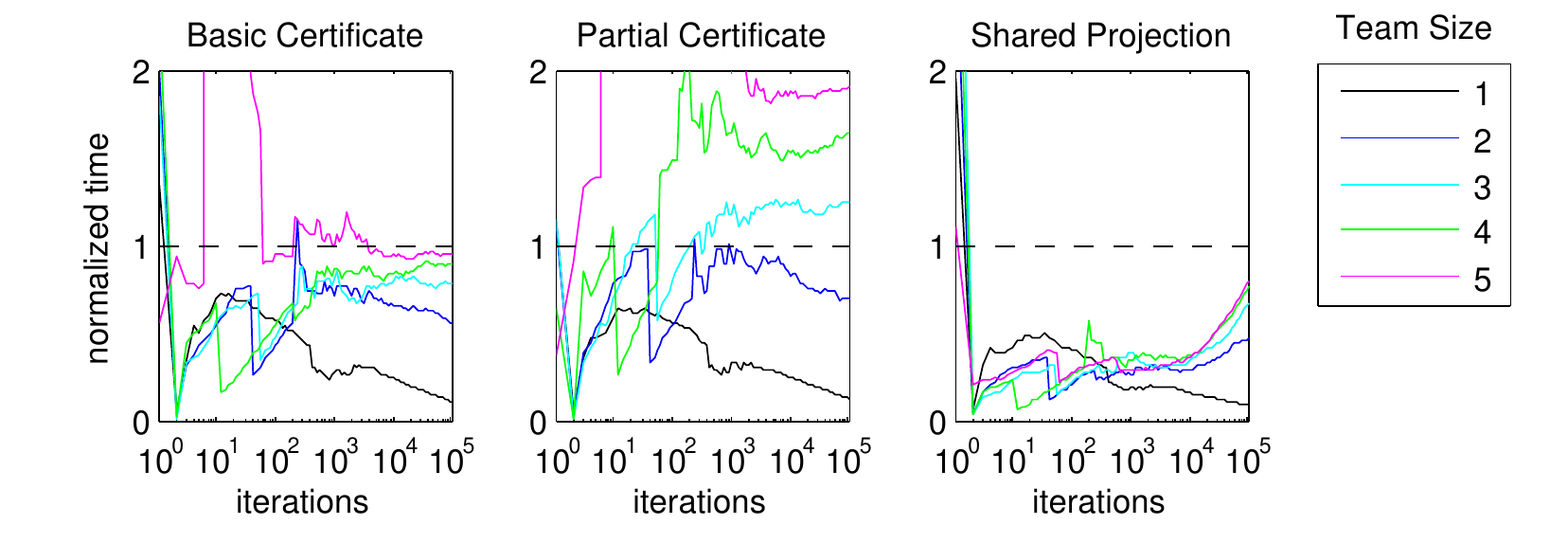}

    \caption[]{Relative runtime of certificate methods vs. normal collision checking (mean over 20 trials), points below the dotted line are desired. The RRT* algorithm is used. Top to Bottom depict the RRT* algorithm with collision difficulty levels of $1$ (raw data), $10^2$, and $10^4$, respectively.}
    \label{fig:time_star}

\end{figure}

Figures~\ref{fig:time} and \ref{fig:time_star} show the corresponding runtime---normalized by the  runtime of a standard implementation of RRT or RRT* that does not use any certificate method (averaged over 20 trials), respectively. The top-most sub-figure in \ref{fig:time} and \ref{fig:time_star} show the raw results averaged over 20 trials. 
In order to evaluate how our certificate method performs as collision checking becomes more difficult, the bottom two sub-figures show what happens when collision checking time is increased by a factor of $10^2$ and $10^4$, respectively. These graphs are created by recording the cumulative time spent within the standard collision checking call vs. the total runtime, and then increasing the collision checking time by the desired multiple while holding the non-collision checking time constant. This is only an approximation to what might be expected in practice for difficult collision checking scenarios; however, we believe that it provides useful insight into how each method should be expected to perform as collision checking becomes more difficult.

\section{Results and Conclusions}

With respect to the proportion of collision checks that are avoided, {\it Basic Certificate} suffers from a curse of dimensionality that limits its usefulness for team sizes greater than $2$ (Figure~\ref{fig:ratio}). {\it Shared Projection} provides significant runtime reductions for all team sizes and difficulty levels (Figure~\ref{fig:time}); however, there is eventually a graph size for which using certificates becomes more expensive than a traditional collision check. We note that a similar result was also observed in the single robot version of this work; however, the crossover point was located orders of magnitude latter. 
We believe that this happens sooner for {\it Shared Projection} due to the fact that $R$ nodes must be added to the secondary kd-tree per iteration. In practice, this can be dealt with by switching to standard collision checks once the size of the kd-tree makes using certificates (e.g., going back up the kd-tree after a kd-search search) disadvantageous. The advantages of using {\it shared projection} last longer as collision checking becomes more difficult---therefore, we expect it to be most useful in the particular scenarios most in need of collision checking efficiency.

Our main results can be summarized as follows:
\begin{itemize}
\item Either {\it Partial Certificate} or {\it Shared Projection} should be used instead of {\it Basic Certificate}.
\item {\it Shared Projection} should be used when the team size is relatively large (e.g., $R>2$), and the number of nodes is sufficiently small that the benefits of reduced collision checking outweigh the extra kd-tree overhead (e.g., when collision checking is relatively expensive).
\item {Partial Certificates} will provide moderate improvements when team size is small ($R \leq 2$) and collision checking is relatively inexpensive. 
\end{itemize}

\bibliographystyle{plain}
\bibliography{publications}

%


\end{document}